\documentclass[conference]{IEEEtran}
\IEEEoverridecommandlockouts

\usepackage{cite}
\usepackage{amsmath,amssymb,amsfonts}
\usepackage{graphicx}
\usepackage{textcomp}
\usepackage{booktabs}
\usepackage{subfig}
\usepackage{algorithm} 
\usepackage{algpseudocode} 
\usepackage{multirow}
\usepackage{colortbl} 
\usepackage{xcolor}
\def\BibTeX{{\rm B\kern-.05em{\sc i\kern-.025em b}\kern-.08em
    T\kern-.1667em\lower.7ex\hbox{E}\kern-.125emX}}
\begin{document}

\title{Mixed-Precision Graph Neural Quantization for Low Bit  Large Language Models 
}


\vspace{3pt}
\author{Wanlong Liu\textsuperscript{\rm 1}$^\dagger$,  
Yichen Xiao\textsuperscript{\rm 1}$^\dagger$, \thanks{$\dagger$ These authors contributed equally to this work.}
Dingyi Zeng\textsuperscript{\rm 1},  
Hongyang Zhao\textsuperscript{\rm 1},  
Wenyu Chen\textsuperscript{\rm 1}, 
Malu Zhang\textsuperscript{\rm 1}*\thanks{*Corresponding Author} \\

\vspace{3pt}

\textit{School of Computer Science and Engineering,}\textit{ University of Electronic Science and Technology of China\textsuperscript{\rm 1}} \\
}  

\vspace{3pt}

\maketitle

\begin{abstract}
Post-Training Quantization (PTQ) is pivotal for deploying large language models (LLMs) within resource-limited settings by significantly reducing resource demands. 
However, existing PTQ strategies underperform at low bit levels ($< \text{3 bits}$) due to the significant difference between the quantized and original weights. 
To enhance the quantization performance at low bit widths, we introduce a \textbf{M}ixed-precision \textbf{G}raph Neural  PTQ (MG-PTQ) approach, employing a graph neural network (GNN) module to capture dependencies among weights and adaptively assign quantization bit-widths.
Through the information propagation of the GNN module, our method more effectively captures dependencies among target weights, leading to a more accurate assessment of weight importance and optimized allocation of quantization strategies.
Extensive experiments on the WikiText2 and C4 datasets demonstrate that our MG-PTQ method outperforms previous state-of-the-art PTQ method GPTQ, setting new benchmarks for quantization performance under low-bit  ($<\text{3 bits}$) conditions.    

\end{abstract}


\section{Introduction}
Recently, large language models (LLMs) such as the GPT~\cite{achiam2023gpt} and LLaMA~\cite{touvron2023open} series have achieved remarkable performance on various natural language benchmarks~\cite{ yu2025application, bai2024mt, li2024graphreader,  xu2025hybrid, wang2016information, hendrycks2020measuring, wu2024conceptmath, yu2024optimization, xu2025enhancing}. 
However, their deployment is challenging due to large parameter sizes and high memory demands.
For example, one LLaMA2-70B model~\cite{touvron2023open}, with 70 billion parameters, requires 150GB of storage in half-precision format, necessitating at least two A100 GPUs with 80GB each for inference~\cite{huang2024good}. 
Such substantial resource requirements highlight the urgent need for efficient light-weighting techniques to reduce the deployment constraints in resource-limited settings. 

Model quantization has emerged as a highly effective technology among various lightweight methods for compressing LLMs~\cite{dettmers2023spqr}. The primary quantization techniques primarily fall into Quantization-Aware Training (QAT)~\cite{tailor2020degree, ni2024earnings} and Post-Training Quantization (PTQ)~\cite{frantar2022gptq, huang2024billm, shang2023pb}.  
Compared to QAT, PTQ simplifies computations by eliminating backpropagation, which speeds up quantization and increases its practicality~\cite{huang2024billm, huang2024slim}.

Recent mainstream PTQ methods typically use Cholesky matrix decomposition~\cite{higham1990analysis} of the second-order Hessian matrix to measure the importance of weights in LLMs, thereby guiding the quantization strategy. 
However, these methods still have significant performance degradation under low bit quantization ($<\text{3 bits}$). 
This degradation is primarily due to the significant quantization error between the quantized and original weights~\cite{huang2024billm} at low bit levels, posing a major challenge for current PTQ approaches.

To minimize quantization error, it's essential to accurately assess the importance of model weights, assigning higher bit-widths to critical ones and lower bit-widths to less critical ones, to preserve overall model performance.
To achieve this, we propose a \textbf{M}ixed-precision \textbf{G}raph Neural PTQ (MG-PTQ) method. This approach uses a graph neural network (GNN) to perceive dependencies between weights and \textbf{adaptively assign quantization bit-widths} based on their importance, while \textbf{maintaining a controllable average bit-width}. 
Through the information propagation of the GNN, the model can better perceive the dependencies between weights, enabling a more accurate evaluation of their importance and allowing for more optimal allocation of quantization strategies.

In our MG-PTQ approach, the constructed feature graph represents each column of target weights as a node, with the second-order Hessian matrix serving as the weighted adjacent matrix and target weight values as node features.  During training, the GNN module's objective is to minimize quantization error, while being constrained by a penalty on average bit-width to ensure a controllable average bit-width. To address the issue of discontinuous gradients caused by the discrete bit-widths output by the GNN module, we propose an Approximate Gradient strategy, which enables the GNN model to optimize its parameters by minimizing quantization error.



In summary, our main contributions are as follows:
\begin{itemize} 

\item We propose a Mixed-precision Graph Neural PTQ method that adaptively perceives weight importance and allocates quantization bit widths. 
To our knowledge, this is the first work to leverage GNNs for adaptive quantization of LLMs.

\item Our method allows for customized quantization objectives by leveraging the GNN framework. This adaptability allows us to control the target bit-widths for mixed-precision quantization.

\item Extensive experiments on WikiText2~\cite{merity2016pointer} and C4~\cite{raffel2020exploring} datasets demonstrate that our proposed architecture achieves state-of-the-art performance in low-bit scenarios. Efficiency analysis and ablation experiments show that our method is both computationally efficient and adaptable, outperforming existing PTQ approaches.

\end{itemize}


\section{Related Works}

\subsection{Large Language Model Quantization}

Quantization refers to the process of converting high-precision weights and activations of a model into lower precision. For large language models, it serves as an effective method to reduce computational resources and memory requirements, significantly improving inference efficiency and energy performance. Quantization can be categorized into two main types: Quantization-Aware Training (QAT) and Post-Training Quantization (PTQ). QAT preserves model performance through quantization-aware training. For example, LLM-QAT~\cite{liu2023llm} addressed data barrier issues in QAT training through data-free distillation. QLoRA~\cite{dettmers2024qlora} introduces an efficient finetuning method that significantly reduces memory usage. However, retraining large models is costly, QAT methods require substantial GPU time. 


PTQ then has become a more efficient alternative, capable of scaling to large models~\cite{li2021brecq, shang2023pb, frantar2022gptq, yao2022zeroquant, huang2024billm, lin2024awq, dettmers2023spqr}. BRECQ~\cite{li2021brecq} improves quantization accuracy by introducing additional grouping labels for custom quantization blocks. PB-LLM~\cite{shang2023pb} presents a partially-binarized approach that selectively stores salient weights in higher precision, enabling low-bit quantization of large language models. GPTQ~\cite{frantar2022gptq} offers a highly efficient one-shot quantization method using approximate second-order information, reducing the size of large GPT models to 3-4 bits per weight with minimal accuracy loss. Similarly, BiLLM~\cite{huang2024billm} mitigates quantization loss by selecting important weights and applying a binary residual approximation strategy. It also employs an optimal grouping method to ensure high-precision inference while maintaining strong time efficiency. However, these approaches still suffer from notable performance degradation under low bit quantization.

\subsection{Graph Neural Networks}
\label{sec:NGraph Neural Networks}

The concept of Graph Neural Networks (GNNs) can be traced back to foundational works such as \cite{sperduti1997supervised,zeng2023substructure}, with the use of neural networks for representing graph data and extracting features having a long-standing history of development, as highlighted in studies like \cite{gori2005new,scarselli2008graph,zeng2022simple,zeng2023rethinking}. A significant evolution of this idea came with the introduction of Graph Convolutional Networks \cite{DBLP:conf/iclr/KipfW17}, which played a pivotal role in semi-supervised node classification and greatly advanced the field. Over time, GNN applications have expanded into unsupervised tasks such as graph contrastive learning and node clustering \cite{DCRN,yue2022survey,liu2022simple,liu2023hsan,chen2023attribute,zeng2023substructure}.
The existing quantization techniques for LLMs fail to well capture the importance of target weights, resulting in significant quantization error when quantizing to lower bit levels. 
\begin{figure}[ht] 
    \centering 
    \includegraphics[scale=0.23]{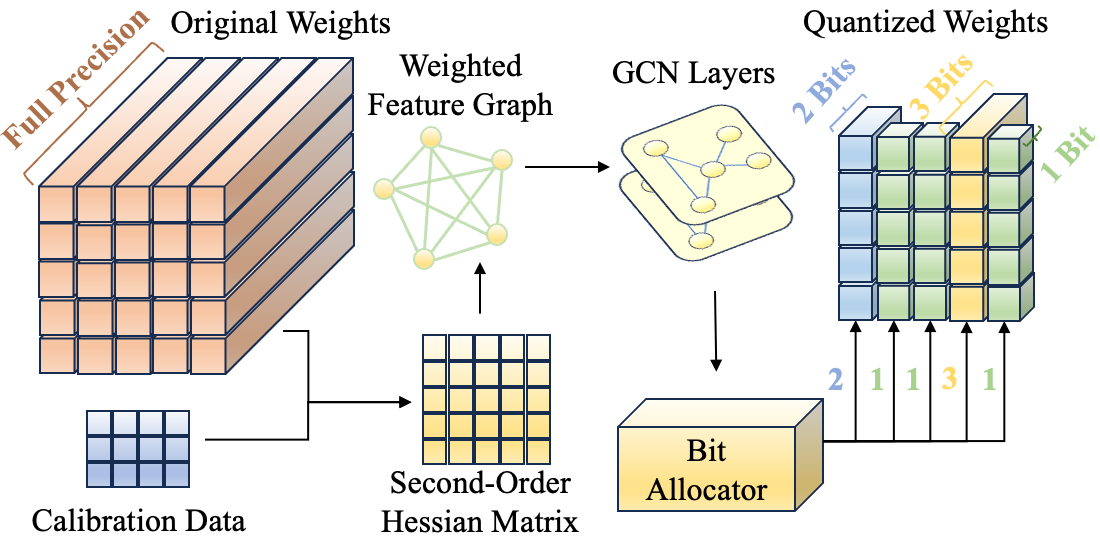}
    \centering
    \caption{The overall architecture of MG-PTQ model.}
    \label{architecture}
\end{figure}

\section{Method}
\label{sec:method}
In this section, we first introduce the preliminary knowledge on quantization, covering both multi-bit and binary quantization techniques. Following this, we introduce our Graph-based Mixed-precision PTQ method. Finally, we describe the training process of our GNN module. The overall architrcture of our MG-PTQ method is shown in Fig.~\ref{architecture}.
\subsection{Preliminaries}
Quantization maps continuous inputs to discrete values. From a nonlinear mapping perspective, continuous values can be quantized into integers by combining multiple binary quantizations. A low-bit quantization function is expressed as a sum of unit step functions with specific biases and scaling factors:
\[
y = \sum_{i=1}^n s_i A(\beta (x - b_i)) - o,
\]
where \(x\) is the input, \(y\) the quantized output, \(s_i\) and \(b_i\) the scaling factor and bias for each step, \(A\) the unit step function, and \(o\) the offset for zero-centering.
Two approaches are common: uniform and non-uniform quantization. Uniform quantization keeps equal intervals between levels, achieved by using identical \(s_i\) and uniformly distributed \(b_i\). Non-uniform quantization adjusts intervals based on input distribution, with finer levels in regions of significant variation to reduce quantization error.
For binary quantization, the simplified equation is:
\begin{equation}
Q_x = \alpha \text{sign}(x),
\end{equation}
where \( \alpha \) is a scaling factor enhancing the binary output, and \(\text{sign}(x)\) function outputs \(+1\) for \(x \geq 0\) and \(-1\) for \(x < 0\). 
This simplifies computations by reducing complex multiplications within neural networks to simpler operations.

\begin{algorithm}

\caption{Pipeline of MG-PTQ}
\begin{algorithmic}[1] 
\State \textbf{Input:} \(\mathbf{W}^{{t}} \in \mathbb{R}^{d_\text{row} \times d_\text{col}}\) - target weight, \(\mathbf{X}_F \in \mathbb{R}^{m \times d_\text{col}}\) - calibration data, \(\beta\) - block size, \(\lambda\) - Hessian regularizer
\State \textbf{Output:} \(\mathbf{B}\) - binarized weights
\State \(\mathbf{H}^c := \text{Cholesky}\left((2\mathbf{X}_F^T \mathbf{X}_F + \lambda \mathbf{I})^{-1}\right)\) // Hessian matrix 
\State \( \mathbf{W}^t := \text{Preprocess}(\mathbf{W}^t) \) // Apply transpose, padding.
\State \( \mathbf{X}_G^{(0)} := \frac{1}{k} \sum_{i=1}^k \mathbf{W}^t_{ :, i, :} \) 
\State \(\mathbf{X}_G^{(L)} := \text{GCN}(\mathbf{H}^{c}, \mathbf{X}_G^{(0)})\) // Graph Perceptual Module
\State \(B^{\text{width}} := \arg\max(\text{softmax}(\text{FFNN}(\mathbf{X}_G^{(L)})))\)
\State \(\mathbf{B} := 0_{{d_\text{row} \times d_\text{col}}}\)
\For{\(b = 0, \beta, 2\beta, \ldots, N\)} 
    \State \( \mathbf{W}^b := \mathbf{W}_{:,b:b+\beta} \)
    \For{\(t = 1\) to \(t_\text{max}\)} // \(t_\text{max}\) denotes the max bit-width
            \State \( \mathbf{B}_t := t\text{-bit quantization}(\{\mathbf{W}^b_{i, :} \mid B^{\text{width}}_i = t\}) \)
    \EndFor
    \State \( \mathbf{B}_{:,b:b+\beta} := \mathbf{B}_1 + \mathbf{B}_2 + ... + \mathbf{B}_{t_\text{max}}\)
    \State \( \mathbf{E} := (\mathbf{W}_{:,b:b+\beta} - \mathbf{B}_{:,b:b+\beta}) / \mathbf{H}^{c}_{b:b+\beta,b:b+\beta} \)
\State \( \mathbf{W}_{:,b:b+\beta} := \mathbf{W}_{:,b:b+\beta} - \mathbf{E} \cdot \mathbf{H}^{c}_{b:b+\beta,b:b+\beta} \) // {Block-wise Output Compensation (OBC)}
\EndFor
\State \Return \(\mathbf{B}\)
\end{algorithmic}
\label{algorithm1}
\end{algorithm}

\subsection{Mixed-precision Graph Neural  PTQ}

Our MG-PTQ method leverages a GNN to evaluate weight importance and adaptively assign quantization bit-widths.  
It consists of two modules: a graph-based perceptual module that captures dependencies between weights, and a bit-width allocator that allocates quantization bit-widths based on the importance of the weights.

\subsubsection{Graph Perceptual Module}
To assess the importance of weights, we construct a feature graph where each column of weights is represented as a node. 
We use the Cholesky decomposition~\cite{higham1990analysis} of the second-order Hessian matrix 
$\mathbf{H}^c \in \mathbb{R}^{d_\text{col} \times d_\text{col}}$ as the weighted adjacency matrix\footnote{According to~\cite{frantar2022gptq}, the Cholesky reformulation improves numerical stability and removes the need for Hessian updates, reducing computation.}, with the weight values serving as node features.
Specifically, given the calibration data $\mathbf{X_F} \in \mathbb{R}^{m \times d_\text{col}}$, \(\mathbf{H}^c\) can be calculated as follows:

\begin{equation}
\setlength\abovedisplayskip{3pt plus 3pt minus 7pt}
\setlength\belowdisplayskip{3pt plus 3pt minus 7pt}
\mathbf{H}^c = \text{Cholesky}\left((2\mathbf{X}_F^{T} \ \mathbf{X}_F  + \lambda \mathbf{I})^{-1}\right),
\end{equation}
where $\text{Cholesky}$ denotes the Cholesky decomposition.
Given the target weight \(\mathbf{W}^{{t}} \in \mathbb{R}^{d_{\text{row}} \times d_{\text{col}}}\), we first pad and reshape it to the nearest multiple of \(d_{\text{gnn}}\), resulting in a new shape of \(\mathbb{R}^{d_{\text{col}} \times k \times d_{\text{gnn}}}\). We then take the mean along the first dimension and obtain the input feature for the GNN module \(\mathbf{X_{G}}^{(0)} \in \mathbb{R}^{d_{\text{col}} \times d_{\text{gnn}}}\) as follows:

\begin{equation}
\setlength\abovedisplayskip{3pt plus 3pt minus 7pt}
\setlength\belowdisplayskip{3pt plus 3pt minus 7pt}
\mathbf{X_{G}}^{(0)} = \frac{1}{k} \sum_{i=1}^{k} \mathbf{W}^{{t}}_{:,i,:}.
\end{equation}

We utilize a 2-layer GCN~\cite{kipf2016semi} to process the constructed feature graph, employing its message-passing mechanism to capture the dependencies between weights as follows:

\begin{equation}
\setlength\abovedisplayskip{3pt plus 3pt minus 7pt}
\setlength\belowdisplayskip{3pt plus 3pt minus 7pt}
\mathbf{X_{G}}^{(l+1)} = \sigma\left(\mathbf{H}^c \mathbf{X_{G}}^{(l)} \mathbf{W}^{(l)}\right),
\end{equation}
where \(\mathbf{X_{G}}^{(l)}\) represents the input feature for the \(l\)-th layer of the GNN, and \(\mathbf{W}^{(l)}\) represents the weight of the \(l\)-th GNN layer. $\sigma$ denotes the activation function, such as ReLU.

\subsubsection{Bit-Width Allocator}   The bit-width allocator is a feed-forward neural network (FFNN) classifier that assigns quantization bit-widths to each column of the target weight based on the features from the last layer of the GCN:

\begin{equation}
\setlength\abovedisplayskip{3pt plus 3pt minus 7pt}
\setlength\belowdisplayskip{3pt plus 3pt minus 7pt}
B^{\text{width}} =\arg\max\left(\text{softmax}(\text{FFNN}(\mathbf{X_G}^{(L)}))\right),
\label{eq:argmax}
\end{equation}
where $B^{\text{width}}$ denotes the bit-width assigned to each column of the weights.
The number of classes in the classifier is a hyperparameter that sets the maximum quantization bit-width.

\subsubsection{Quantization}  
Following the GPTQ approach~\cite{frantar2022gptq}, we adopt blockwise quantization, which divides weights into smaller blocks for localized adjustments, balancing model size reduction with performance retention. This strategy effectively minimizes quantization error while maintaining flexibility and precision. The detailed process is shown in Algorithm~\ref{algorithm1}.

\begin{table*}[htb]

\caption{Main experimental results. We compare the performance of quantization at different bis for the baselines and our models on the WikiText2 and C4 datasets. The evaluation metric is perplexity (PPL), with lower values indicating better performance. Here, 1-7b refers to the LLaMA1-7b model, and similarly for other models.}
\centering
\small{
\begin{tabular}{@{}ccccccc|ccccc@{}}
\toprule
                                                        &                                                                                     & \multicolumn{5}{c|}{\textbf{WikiText2}}   & \multicolumn{5}{c}{\textbf{C4}}                                                                                                                                                                                                          \\ \cmidrule(l){3-12} 
\multirow{-2}{*}{\textbf{Method}}                       & \multirow{-2}{*}{\textbf{\begin{tabular}[c]{@{}c@{}}Weight\\ Bits\end{tabular}}} & 1-7b   & 1-13b  & 2-7b  & 2-13b  & 3-8B   & 1-7b                                         & 1-13b                                        & 2-7b                                         & 2-13b                                        & 3-8B                                         \\ \midrule
Full Precision                                          & 16                                                                                  & 5.68   & 5.09   & 5.47  & 4.88   & 5.75   & 7.08                                         & 6.61                                         & 6.97                                         & 6.46                                         & 9.22                                         \\ \midrule
RTN                                                   & 2                                                                                   & 1.9e3  & 781.20 & 4.2e3 & 122.08 & 1.9e3  & 1.0e3                                        & 451.22                                        & 4.9e3                                        & 140.46                                        & 2.5e4                                        \\
AWQ~\cite{lin2306awq}                                                     & 2                                                                                   & 2.6e5  & 2.8e5  & 2.2e5 & 1.2e5  & 1.7e6  & 1.9e5                                        & 2.3e5                                        & 1.7e5                                        & 9.4e4                                        & 2.1e6                                        \\
GPTQ~\cite{frantar2022gptq}                                                    & 2                                                                                   & 152.31 & 50.44  & 60.45 & 49.70  & 210.00 & 101.3                                        & 63.24                                        & 126.12                                       & 59.68                                        & 4.1e4                                        \\



GPTQ~\cite{frantar2022gptq}                                                    & 1                                                                                   & 2.7e5  & 1.1e5  & 1.1e5 & 9.4e3  & -      & 4.4e5                                        & 2.9e5                                        & 1.2e6                                        & 2.7e3                                        & -                                            \\


 \midrule
\rowcolor[HTML]{FFFFFF} 
& 2.5                                                                                   & \textbf{39.53} & \textbf{12.26}  & \textbf{29.64}& \textbf{13.47}  & \textbf{46.38} & \textbf{29.43}                                        & \textbf{10.37}                                        & \textbf{22.85}                                       & \textbf{11.18}                                        & \textbf{48.31}                                       \\

                                                        & 2                                                                                   & \text{130.27} & \text{49.54}  & \text{58.24}& \text{44.24}  & \text{102.82} & \text{96.70}                                        & \text{31.24}                                        & \text{118.61}                                       & \text{28.18}                                        & \text{181.89}                                       \\
\rowcolor[HTML]{FFFFFF} 
                                                        & 1.8                                                                                 & 128.16 & 82.14  & 180.46& 74.86  & 251.28 & 224.66                                     & 75.71                                        & 145.61                                      & 67.25                                        & 278.64                                     \\
\rowcolor[HTML]{FFFFFF} 
\multirow{-4}{*}
{\cellcolor[HTML]{FFFFFF}MG-PTQ (Ours)} & 1.6                                                                                 & 263.18 & 110.02 & 280.20 & 120.12 & 460.42 & 261.42                                      & 98.65                                        & 296.84                                       & 82.82                                        & 480.64                                      \\ 
\bottomrule
\end{tabular}
\label{tab:main_results}}
\end{table*}

\subsection{Training of the GNN Module}
To train our GCN module, we set two optimization objectives: (1) quantization error and (2) average bit-width. Note that during the training process, only the GNN module is trained while the parameters of the LLMs are frozen.

\subsubsection{Quantization Error}  
We utilize the quantization error, denoted as $\mathbf{E}$, from line 15 in Algorithm~\ref{algorithm1} and sum across all blocks to define our optimization objective $L_{\text{quant}}$. 

However, due to the \textit{argmax} operation in Equation~\ref{eq:argmax} which leads to discontinuous gradients, we employ an Approximate Gradient strategy. Specifically, we employ the \textit{Gumbel-Softmax} operation~\cite{jang2016categorical} as a replacement for the standard \textit{softmax} and \textit{argmax} operations, allowing us to convert predicted probabilities into discrete quantization bits effectively, akin to the results of \textit{argmax}. 
We employ the approximation strategy only during the training phase. During inference, we use the \textit{argmax} function to determine the quantization bit-widths of the target weights.

\subsubsection{Average Bit-width Constrain}  
To ensure that the average bit-width of quantization is controllable, we introduce a penalty on the average bit-width, denoted as $L_\text{bit}$. Specifically, $L_\text{bit}$ represents the Mean Squared Error (MSE) Loss between the target bit-width and the average quantization bit-width achieved by the GNN model.  
Throughout the training process,  $L_\text{bit}$ ensures that the model's average quantization bit-width closely aligns with the target.

The final loss function for our method can be defined as:
\begin{equation}
L = L_{\text{quant}} + \alpha L_{\text{bit}}.
\end{equation}
where \( L_{\text{quant}} \) represents the task-related loss of the model, \( L_{\text{bit}} \) is the mean squared error loss between the target and actual average quantization bit-widths, and \( \alpha \) is a hyperparameter that balances the importance.
During training, we use weights from all layers of the inference model as training data to optimize our total loss function. We set the gradient accumulation steps to 4, meaning that parameters are updated after every four weight quantization. Finally, the trained GNN model is used to allocate the quantization weights of the inference model and calculate the average bit width.

\section{Experiments}
\label{sec:Experiments}

\subsection{Experimental Setup}
\subsubsection{Models \& Metrics \& Datasets}
We implement our approach on the LLaMA model families~\cite{touvron2023open}, including LLaMA-7b, LLaMA-13b, LLaMA2-7b, LLaMA2-13b and LLaMA3-8b models. Perplexity is employed as the evaluation metric, which is widely recognized for its challenge in reliably assessing LLM performance, especially in research centered on network compression~\cite{huang2024billm}.  Our evaluation experiments utilize the WikiText2~\cite{merity2016pointer} and C4~\cite{raffel2020exploring} datasets.

\subsubsection{Baselines} 
We compare our MG-PTQ with various quantization baselines that do not require additional training or fine-tuning, includeing classic PTQ methods such as vanilla round-to-nearest (RTN) and GPTQ~\cite{frantar2022gptq}, along with AWQ~\cite{lin2306awq}.

\subsubsection{Implement Details}

We set 50 training epochs, employ 4 gradient accumulation steps, and apply a learning rate of 0.001. The GCN module is set with an input dimension of 512 and a maximum bit width of 4. 
Additionally, we set the $\alpha$ to 1, utilize the AdamW optimizer, and a block size of 128 following~\cite{achiam2023gpt}.



\subsection{Main Results}
In this section, we conduct experiments on the WikiText2 and C4 datasets within the LLaMA family, reporting the perplexity after quantization at various bit levels. (1) As shown in Table~\ref{tab:main_results}, traditional quantization methods like RTN and AWQ exhibit extremely poor performance at 1-bit and 2-bit levels, highlighting the challenges faced by existing PTQ methods in low-bit scenarios. (2) Compared to previous state-of-the-art PTQ method GPTQ, our MG-PTQ outperforms it at 2 bit. This demonstrates the effectiveness of our graph perception module in more accurately assessing the importance of weights and allocating quantization bits.
\subsection{Analysis}
\subsubsection{Ablation Study}
\begin{figure}[tbp]
    \centering
    \subfloat[\label{fig2:a}Ablation Study]{
    \includegraphics[width=0.5\linewidth]{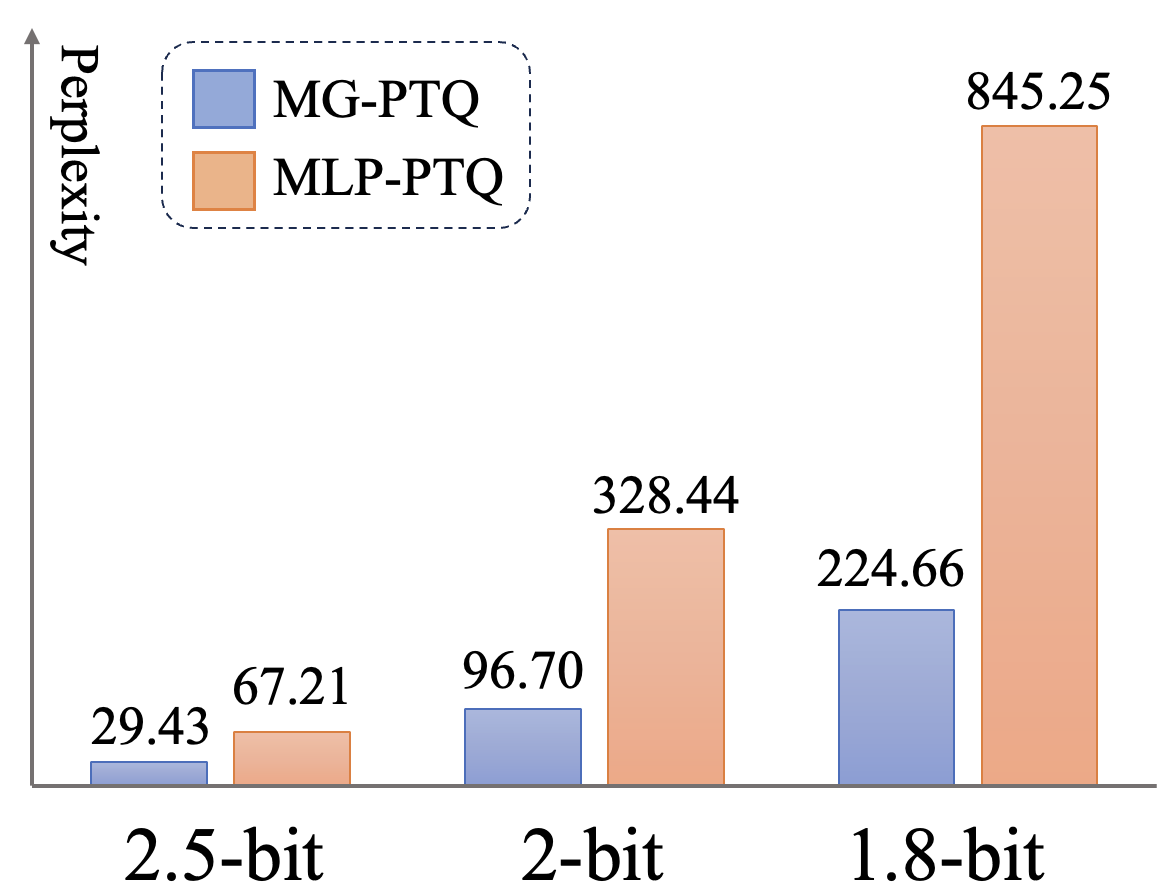}
    }
    \
    \subfloat[\label{fig:b} Efficiency Analysis]{
    \includegraphics[width=0.35\linewidth]{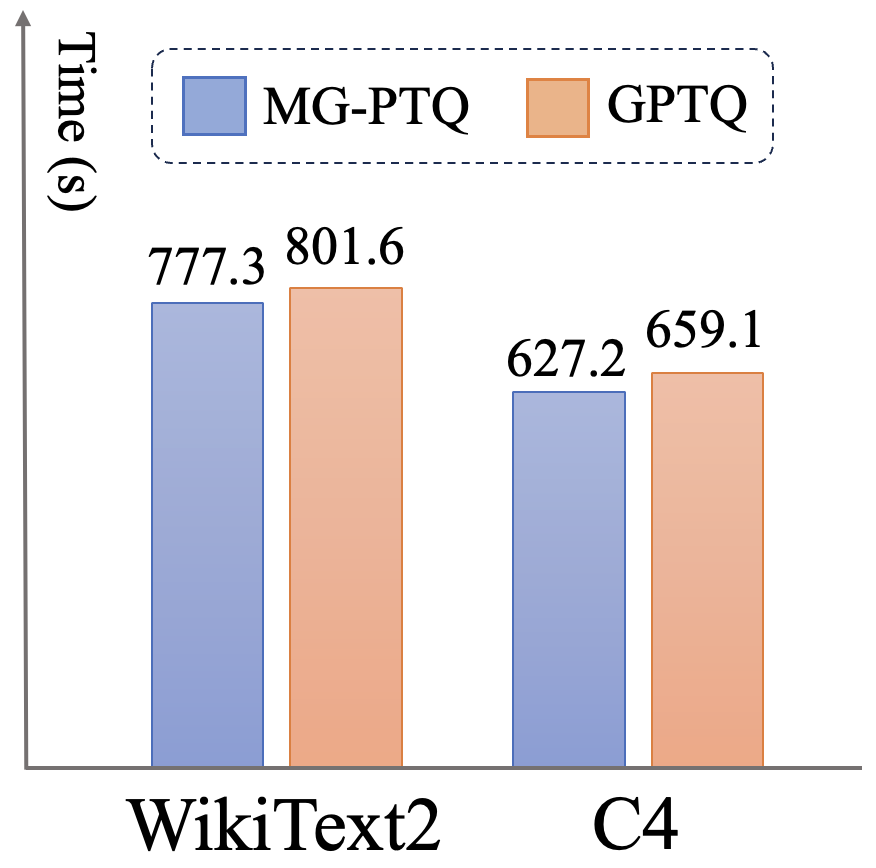}
    }

    \caption{Further experimental analysis. Sub-figure (a) presents the ablation study of LLaMA-7b model on C4 dataset, across different quantization bit Depths. And Sub-figure (b) shows the efficiency analysis, where quantization time of LLaMA-7b model is tested across different quantization strategies.}
    \label{fig:fig2}
\end{figure}

To evaluate the effectiveness of our graph perception module in weight importance perception, we perform an ablation study. Specifically, we replace the GCN module with an MLP (referred to as \textbf{MLP-PTQ}) and compare the quantization performance of the LLaMA-7b model at 1.8, 2.0, and 2.5 bits on C4 dataset. For the MLP, the input feature is the second-order Hessian matrix. As shown in Fig.~\ref{fig:fig2} (a), replacing the GCN module with an MLP results in a significant drop in performance, highlighting the crucial role of the GCN module in assessing weight importance.

\subsubsection{Effiency Analysis}
Our MG-PTQ method can adaptively allocate a controllable target bit-width while maintaining efficient quantization. We measure and compare the quantization times of MG-PTQ and MLP-PTQ applied to LLaMA-7b on one RTX 4090 GPU. As shown in Fig.~\ref{fig:fig2} (b), MG-PTQ achieves nearly the same quantization time as GPTQ, but it surpasses GPTQ in low-bit quantization performance while providing precise control over bit-width allocation.
\section{Conclusion}
\label{sec:Conclusion}
In this paper, to tackle the issue of poor quantization performance at low bit levels seen in PTQ methods, we propose the Mixed-precision Graph Neural  PTQ (MG-PTQ) approach. This method utilizes a graph neural network to perceive dependencies among weights and adaptively assigns quantization bit-widths, effectively enhancing the focus on crucial weight importance. 

\section*{Acknowledgements}
This work was supported in part by the National Natural Science Foundation of China under grant U20B2063, 62220106008, and 62106038, the Sichuan Science and Technology Program under Grant 2024NSFTD0034 and 2023YFG0259.

\clearpage
\bibliographystyle{IEEEtran}
\bibliography{strings,refs}

\begin{thebibliography}{10}
\providecommand{\url}[1]{#1}
\csname url@samestyle\endcsname
\providecommand{\newblock}{\relax}
\providecommand{\bibinfo}[2]{#2}
\providecommand{\BIBentrySTDinterwordspacing}{\spaceskip=0pt\relax}
\providecommand{\BIBentryALTinterwordstretchfactor}{4}
\providecommand{\BIBentryALTinterwordspacing}{\spaceskip=\fontdimen2\font plus
\BIBentryALTinterwordstretchfactor\fontdimen3\font minus \fontdimen4\font\relax}
\providecommand{\BIBforeignlanguage}[2]{{%
\expandafter\ifx\csname l@#1\endcsname\relax
\typeout{** WARNING: IEEEtran.bst: No hyphenation pattern has been}%
\typeout{** loaded for the language `#1'. Using the pattern for}%
\typeout{** the default language instead.}%
\else
\language=\csname l@#1\endcsname
\fi
#2}}
\providecommand{\BIBdecl}{\relax}
\BIBdecl

\bibitem{achiam2023gpt}
J.~Achiam, S.~Adler, S.~Agarwal, L.~Ahmad, I.~Akkaya, F.~L. Aleman, D.~Almeida, J.~Altenschmidt, S.~Altman, S.~Anadkat \emph{et~al.}, ``Gpt-4 technical report,'' \emph{arXiv preprint arXiv:2303.08774}, 2023.

\bibitem{touvron2023open}
H.~Touvron, T.~Lavril, G.~Izacard, X.~Martinet, M.~Lachaux, T.~Lacroix, B.~Rozi{\`e}re, N.~Goyal, E.~Hambro, F.~Azhar \emph{et~al.}, ``Open and efficient foundation language models,'' \emph{Preprint at arXiv. https://doi. org/10.48550/arXiv}, vol. 2302, 2023.

\bibitem{yu2025application}
P.~Yu, Z.~Xu, J.~Wang, and X.~Xu, ``The application of large language models in recommendation systems,'' \emph{arXiv preprint arXiv:2501.02178}, 2025.

\bibitem{bai2024mt}
G.~Bai, J.~Liu, X.~Bu, Y.~He, J.~Liu, Z.~Zhou, Z.~Lin, W.~Su, T.~Ge, B.~Zheng \emph{et~al.}, ``Mt-bench-101: A fine-grained benchmark for evaluating large language models in multi-turn dialogues,'' \emph{arXiv preprint arXiv:2402.14762}, 2024.

\bibitem{li2024graphreader}
S.~Li, Y.~He, H.~Guo, X.~Bu, G.~Bai, J.~Liu, J.~Liu, X.~Qu, Y.~Li, W.~Ouyang \emph{et~al.}, ``Graphreader: Building graph-based agent to enhance long-context abilities of large language models,'' \emph{arXiv preprint arXiv:2406.14550}, 2024.

\bibitem{xu2025hybrid}
X.~Xu, P.~Yu, Z.~Xu, and J.~Wang, ``A hybrid attention framework for fake news detection with large language models,'' \emph{arXiv preprint arXiv:2501.11967}, 2025.

\bibitem{wang2016information}
D.~Wang, \emph{Information Science and Electronic Engineering: Proceedings of the 3rd International Conference of Electronic Engineering and Information Science (ICEEIS 2016), January 4-5, 2016, Harbin, China}.\hskip 1em plus 0.5em minus 0.4em\relax CRC Press, 2016.

\bibitem{hendrycks2020measuring}
D.~Hendrycks, C.~Burns, S.~Basart, A.~Zou, M.~Mazeika, D.~Song, and J.~Steinhardt, ``Measuring massive multitask language understanding,'' in \emph{9th International Conference on Learning Representations, {ICLR} 2021, Virtual Event, Austria, May 3-7, 2021}.\hskip 1em plus 0.5em minus 0.4em\relax OpenReview.net, 2021.

\bibitem{wu2024conceptmath}
Y.~Wu, J.~Liu, X.~Bu, J.~Liu, Z.~Zhou, Y.~Zhang, C.~Zhang, Z.~Bai, H.~Chen, T.~Ge \emph{et~al.}, ``Conceptmath: A bilingual concept-wise benchmark for measuring mathematical reasoning of large language models,'' \emph{arXiv preprint arXiv:2402.14660}, 2024.

\bibitem{yu2024optimization}
P.~Yu, J.~Yi, T.~Huang, Z.~Xu, and X.~Xu, ``Optimization of transformer heart disease prediction model based on particle swarm optimization algorithm,'' \emph{arXiv preprint arXiv:2412.02801}, 2024.

\bibitem{xu2025enhancing}
X.~Xu, Z.~Xu, P.~Yu, and J.~Wang, ``Enhancing user intent for recommendation systems via large language models,'' \emph{arXiv preprint arXiv:2501.10871}, 2025.

\bibitem{huang2024good}
W.~Huang, X.~Ma, H.~Qin, X.~Zheng, C.~Lv, H.~Chen, J.~Luo, X.~Qi, X.~Liu, and M.~Magno, ``How good are low-bit quantized llama3 models? an empirical study,'' \emph{CoRR}, vol. abs/2404.14047, 2024.

\bibitem{dettmers2023spqr}
T.~Dettmers, R.~Svirschevski, V.~Egiazarian, D.~Kuznedelev, E.~Frantar, S.~Ashkboos, A.~Borzunov, T.~Hoefler, and D.~Alistarh, ``Spqr: {A} sparse-quantized representation for near-lossless {LLM} weight compression,'' in \emph{The Twelfth International Conference on Learning Representations, {ICLR} 2024, Vienna, Austria, May 7-11, 2024}.\hskip 1em plus 0.5em minus 0.4em\relax OpenReview.net, 2024.

\bibitem{tailor2020degree}
S.~A. Tailor, J.~Fern{\'{a}}ndez{-}Marqu{\'{e}}s, and N.~D. Lane, ``Degree-quant: Quantization-aware training for graph neural networks,'' in \emph{9th International Conference on Learning Representations, {ICLR} 2021, Virtual Event, Austria, May 3-7, 2021}.\hskip 1em plus 0.5em minus 0.4em\relax OpenReview.net, 2021.

\bibitem{ni2024earnings}
H.~Ni, S.~Meng, X.~Chen, Z.~Zhao, A.~Chen, P.~Li, S.~Zhang, Q.~Yin, Y.~Wang, and Y.~Chan, ``Harnessing earnings reports for stock predictions: A qlora-enhanced llm approach,'' \emph{arXiv preprint arXiv:2408.06634}, 2024.

\bibitem{frantar2022gptq}
E.~Frantar, S.~Ashkboos, T.~Hoefler, and D.~Alistarh, ``{GPTQ:} accurate post-training quantization for generative pre-trained transformers,'' \emph{CoRR}, vol. abs/2210.17323, 2022.

\bibitem{huang2024billm}
W.~Huang, Y.~Liu, H.~Qin, Y.~Li, S.~Zhang, X.~Liu, M.~Magno, and X.~Qi, ``Billm: Pushing the limit of post-training quantization for llms,'' in \emph{Forty-first International Conference on Machine Learning, {ICML} 2024, Vienna, Austria, July 21-27, 2024}.\hskip 1em plus 0.5em minus 0.4em\relax OpenReview.net, 2024.

\bibitem{shang2023pb}
Z.~Yuan, Y.~Shang, and Z.~Dong, ``{PB-LLM:} partially binarized large language models,'' in \emph{The Twelfth International Conference on Learning Representations, {ICLR} 2024, Vienna, Austria, May 7-11, 2024}.\hskip 1em plus 0.5em minus 0.4em\relax OpenReview.net, 2024.

\bibitem{huang2024slim}
W.~Huang, H.~Qin, Y.~Liu, Y.~Li, X.~Liu, L.~Benini, M.~Magno, and X.~Qi, ``Slim-llm: Salience-driven mixed-precision quantization for large language models,'' \emph{CoRR}, vol. abs/2405.14917, 2024.

\bibitem{higham1990analysis}
N.~J. Higham, ``Analysis of the cholesky decomposition of a semi-definite matrix,'' 1990.

\bibitem{merity2016pointer}
S.~Merity, C.~Xiong, J.~Bradbury, and R.~Socher, ``Pointer sentinel mixture models,'' in \emph{5th International Conference on Learning Representations, {ICLR} 2017, Toulon, France, April 24-26, 2017, Conference Track Proceedings}.\hskip 1em plus 0.5em minus 0.4em\relax OpenReview.net, 2017.

\bibitem{raffel2020exploring}
C.~Raffel, N.~Shazeer, A.~Roberts, K.~Lee, S.~Narang, M.~Matena, Y.~Zhou, W.~Li, and P.~J. Liu, ``Exploring the limits of transfer learning with a unified text-to-text transformer,'' \emph{Journal of machine learning research}, vol.~21, no. 140, pp. 1--67, 2020.

\bibitem{liu2023llm}
Z.~Liu, B.~Oguz, C.~Zhao, E.~Chang, P.~Stock, Y.~Mehdad, Y.~Shi, R.~Krishnamoorthi, and V.~Chandra, ``{LLM-QAT:} data-free quantization aware training for large language models,'' in \emph{Findings of the Association for Computational Linguistics, {ACL} 2024, Bangkok, Thailand and virtual meeting, August 11-16, 2024}.\hskip 1em plus 0.5em minus 0.4em\relax Association for Computational Linguistics, 2024, pp. 467--484.

\bibitem{dettmers2024qlora}
T.~Dettmers, A.~Pagnoni, A.~Holtzman, and L.~Zettlemoyer, ``Qlora: Efficient finetuning of quantized llms,'' \emph{Advances in Neural Information Processing Systems}, vol.~36, 2024.

\bibitem{li2021brecq}
Y.~Li, R.~Gong, X.~Tan, Y.~Yang, P.~Hu, Q.~Zhang, F.~Yu, W.~Wang, and S.~Gu, ``{BRECQ:} pushing the limit of post-training quantization by block reconstruction,'' in \emph{9th International Conference on Learning Representations, {ICLR} 2021, Virtual Event, Austria, May 3-7, 2021}.\hskip 1em plus 0.5em minus 0.4em\relax OpenReview.net, 2021.

\bibitem{yao2022zeroquant}
Z.~Yao, R.~Yazdani~Aminabadi, M.~Zhang, X.~Wu, C.~Li, and Y.~He, ``Zeroquant: Efficient and affordable post-training quantization for large-scale transformers,'' \emph{Advances in Neural Information Processing Systems}, vol.~35, pp. 27\,168--27\,183, 2022.

\bibitem{lin2024awq}
J.~Lin, J.~Tang, H.~Tang, S.~Yang, W.-M. Chen, W.-C. Wang, G.~Xiao, X.~Dang, C.~Gan, and S.~Han, ``Awq: Activation-aware weight quantization for on-device llm compression and acceleration,'' \emph{Proceedings of Machine Learning and Systems}, vol.~6, pp. 87--100, 2024.

\bibitem{sperduti1997supervised}
A.~Sperduti and A.~Starita, ``Supervised neural networks for the classification of structures,'' \emph{IEEE transactions on neural networks}, vol.~8, no.~3, pp. 714--735, 1997.

\bibitem{zeng2023substructure}
D.~Zeng, W.~Liu, W.~Chen, L.~Zhou, M.~Zhang, and H.~Qu, ``Substructure aware graph neural networks,'' in \emph{Proceedings of the AAAI conference on artificial intelligence}, vol.~37, no.~9, 2023, pp. 11\,129--11\,137.

\bibitem{gori2005new}
M.~Gori, G.~Monfardini, and F.~Scarselli, ``A new model for learning in graph domains,'' in \emph{Proceedings. 2005 IEEE international joint conference on neural networks, 2005.}, vol.~2.\hskip 1em plus 0.5em minus 0.4em\relax IEEE, 2005, pp. 729--734.

\bibitem{scarselli2008graph}
F.~Scarselli, M.~Gori, A.~C. Tsoi, M.~Hagenbuchner, and G.~Monfardini, ``The graph neural network model,'' \emph{IEEE transactions on neural networks}, vol.~20, no.~1, pp. 61--80, 2008.

\bibitem{zeng2022simple}
D.~Zeng, L.~Zhou, W.~Liu, H.~Qu, and W.~Chen, ``A simple graph neural network via layer sniffer,'' in \emph{ICASSP 2022-2022 IEEE International Conference on Acoustics, Speech and Signal Processing (ICASSP)}.\hskip 1em plus 0.5em minus 0.4em\relax IEEE, 2022, pp. 5687--5691.

\bibitem{zeng2023rethinking}
D.~Zeng, W.~Chen, W.~Liu, L.~Zhou, and H.~Qu, ``Rethinking random walk in graph representation learning,'' in \emph{ICASSP 2023-2023 IEEE International Conference on Acoustics, Speech and Signal Processing (ICASSP)}.\hskip 1em plus 0.5em minus 0.4em\relax IEEE, 2023, pp. 1--5.

\bibitem{DBLP:conf/iclr/KipfW17}
T.~N. Kipf and M.~Welling, ``Semi-supervised classification with graph convolutional networks,'' in \emph{5th International Conference on Learning Representations, {ICLR} 2017, Toulon, France, April 24-26, 2017, Conference Track Proceedings}.\hskip 1em plus 0.5em minus 0.4em\relax OpenReview.net, 2017.

\bibitem{DCRN}
Y.~Liu, W.~Tu, S.~Zhou, X.~Liu, L.~Song, X.~Yang, and E.~Zhu, ``Deep graph clustering via dual correlation reduction,'' in \emph{Proceedings of the AAAI conference on artificial intelligence}, vol.~36, no.~7, 2022, pp. 7603--7611.

\bibitem{yue2022survey}
Y.~Liu, J.~Xia, S.~Zhou, S.~Wang, X.~Guo, X.~Yang, K.~Liang, W.~Tu, S.~Z. Li, and X.~Liu, ``A survey of deep graph clustering: Taxonomy, challenge, and application,'' \emph{CoRR}, vol. abs/2211.12875, 2022.

\bibitem{liu2022simple}
Y.~Liu, X.~Yang, S.~Zhou, and X.~Liu, ``Simple contrastive graph clustering,'' \emph{arXiv preprint arXiv:2205.07865}, 2022.

\bibitem{liu2023hsan}
Y.~Liu, X.~Yang, S.~Zhou, X.~Liu, Z.~Wang, K.~Liang, W.~Tu, L.~Li, J.~Duan, and C.~Chen, ``Hard sample aware network for contrastive deep graph clustering,'' in \emph{Proceedings of the AAAI conference on artificial intelligence}, vol.~37, no.~7, 2023, pp. 8914--8922.

\bibitem{chen2023attribute}
J.~Chen and G.~Kou, ``Attribute and structure preserving graph contrastive learning,'' in \emph{Proceedings of the AAAI conference on artificial intelligence}, vol.~37, no.~6, 2023, pp. 7024--7032.

\bibitem{kipf2016semi}
T.~N. Kipf and M.~Welling, ``Semi-supervised classification with graph convolutional networks,'' in \emph{5th International Conference on Learning Representations, {ICLR} 2017, Toulon, France, April 24-26, 2017, Conference Track Proceedings}.\hskip 1em plus 0.5em minus 0.4em\relax OpenReview.net, 2017.

\bibitem{lin2306awq}
J.~Lin, J.~Tang, H.~Tang, S.~Yang, X.~Dang, and S.~Han, ``Awq: activationaware weight quantization for llm compression and acceleration. corr, abs/2306.00978, 2023. doi: 10.48550,'' \emph{arXiv preprint ARXIV.2306.00978}.

\bibitem{jang2016categorical}
E.~Jang, S.~Gu, and B.~Poole, ``Categorical reparameterization with gumbel-softmax,'' in \emph{5th International Conference on Learning Representations, {ICLR} 2017, Toulon, France, April 24-26, 2017, Conference Track Proceedings}.\hskip 1em plus 0.5em minus 0.4em\relax OpenReview.net, 2017.

\end{thebibliography}

\end{document}